# Snowy Night-to-Day Translator and Semantic Segmentation Label Similarity for Snow Hazard Indicator

Takato Yasuno[*1] Hiroaki Sugawara[*1], Junichiro Fujii[*1], Ryuto Yoshida[*1]

[*1] Yachiyo Engineering, Co., Ltd., RIIPS.

In 2021, Japan recorded more than three times as much snowfall as usual, so road user maybe come across dangerous situation. The poor visibility caused by snow triggers traffic accidents. For example, 2021 January 19, due to the dry snow and the strong wind speed of 27 m / s, blizzards occurred and the outlook has been ineffective. Because of the whiteout phenomenon, multiple accidents with 17 casualties occurred, and 134 vehicles were stacked up for 10 hours over 1 km. At the night time zone, the temperature drops and the road surface tends to freeze. CCTV images on the road surface have the advantage that we enable to monitor the status of major points at the same time. Road managers are required to make decisions on road closures and snow removal work owing to the road surface conditions even at night. In parallel, they would provide road users to alert for hazardous road surfaces. This paper propose a method to automate a snow hazard indicator that the road surface region is generated from the night snow image using the Conditional GAN, pix2pix. In addition, the road surface and the snow covered ROI are predicted using the semantic segmentation DeepLabv3+ with a backbone MobileNet, and the snow hazard indicator to automatically compute how much the night road surface is covered with snow. We demonstrate several results applied to the cold and snow region in the winter of Japan January 19 to 21 2021, and mention the usefulness of high similarity between snowy night-to-day fake output and real snowy day image for night snow visibility.

## 1. Introduction

### 1.1 Winter Road Safety and Related Works

In Japan, approximately 60 % of land is legally designated as the snow area and cold area, and approximately 20% of people have inhabited regions such as the Hokkaido, North Tohoku, Hokuriku, and Sanin. In the past two decades, approximately 4,000 traffic accidents have occurred per year during winter [NILIM 2003]. The primary reason includes slippery road (91%), reduced ability to see (6%), and rutted road (2%). In winter, following road conditions triggered accidents: iced (50%), dried (29%), wet (11%), and snow (10%). There are several primary hazards of winter driving [Saskatchewan 2019]. They are: 1) poor traction, 2) reduced ability to stop vehicles, 3) the effect of temperature on starting and stopping vehicles, 4) slippery road conditions owing to ice and snow, 5) reduced ability to see and be seen, and 6) jackknifing. Recently, on January 10, 2021, owing to heavy snowfall on the Hokuriku Expressway, approximately 1,000 cars were stuck on the up and down lines between the Fukui and Kanazu interchange. This congestion was caused by a slippery accident. Therefore, road surface monitoring is essential for road safety.

In the past decade, the focus has always been on an intelligent system for monitoring winter road surface conditions via computer vision and machine learning [Loce 2013–Carrillo 2019]. Takahashi et al. studied a prediction method for winter road surfaces corresponding to the change in snow removal levels[Takahashi 2012-'15]. Further, the road surface temperature and condition were estimated based on field observations. However, this method required several observed weather inputs such as temperature, wind velocity, humidity, sky radiation, and sunshine and incurred a higher cost for utilizing multi-mode data. Grabowski et al. proposed an intelligent road sign system including a detector using convolutional neural networks (CNNs) and image processing video cameras for the classification of dry, wet, and snowy roads [Grabowski 2020]. Liang et al. investigated the application of semantic segmentation network customized with dilated convolutions for road surface status recognition [Liang 2019]. The focus has been on performance improvement for road surface recognition; however, an important post-process of providing hazardous information on the winter road surface condition for road managers and users still remains.

### 1.2 Night Road Surface for Snow Hazard Alert

As illustrated in Figure 1, we have previously proposed an application to provide snow hazard index information with snow hazard alerts using live camera images [Yasuno 2021]. If drivers can acquire information on snow levels through live cameras before their travel, they can decide to use snow adaptive tires to avoid tripping, in case of wet and slippery road conditions. The increasing usage of CCTV cameras has opened the possibility of using them to automatically detect hazardous road surfaces and inform drivers through alerts. In winter, we aim to provide a new snow hazard index on the amount of snow covered on the road surface, whose value is from zero to 100. During a heavy snowfall, the road surface region is covered with snow and is not clearly visible on the live camera. Through live image, the background snow excluded with road region would not influence road user for snow hazard. Thus, we need to know the ratio of snow covered on the road.

At the night time zone, the temperature drops and the road surface tends to freeze. The night CCTV images on the road surface have the advantage that we enable to monitor the status of major points at the same time. Road managers are required to make decisions on "road closures" and "snow removal work" owing to the road surface conditions even at night. Simultaneously, they would provide road users to alert for hazardous road surfaces next day early morning. Even if at night

Contact: Takato Yasuno, RIIPS on 3F, 5-20-8, Asakusabashi, Taito-ku, Tokyo, 111-8648, tk-yasuno@yachiyo-eng.co.jp

in order to compute a snow hazard indicator, it is important to recognize the snowy road surface from the night live images.

This study proposes a night-to-day conditional generative adversarial network (GAN) translator from the night to the day using live CCTV images to monitor the road. The night snowy image is translated to the day snowy image using a conditional GAN. Herein, we evaluate the similarity of targeted snow ROI labels between the day real label and the day fake output label using semantic segmentation. Further, the translator are trained and predicted on the snow and cold regions in Japan.

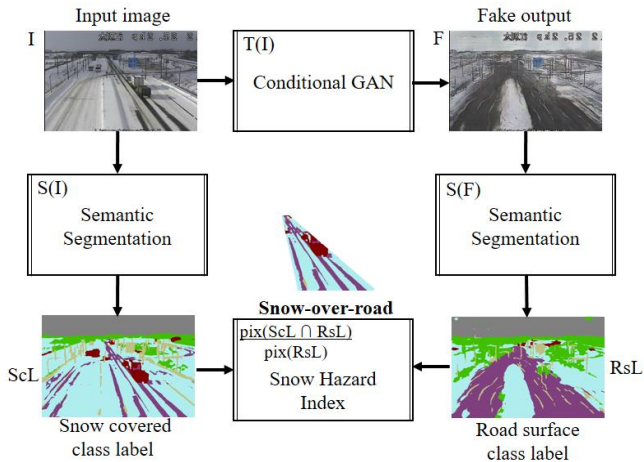

Figure 1: Overview of "previous proposed" pipeline [Yasuno 2021] showing the raw input is translated by the conditional GAN network T. The resulting road surface fake output F is further fed into the semantic segmentation network S to predict a *road surface class label* output RsL. Simultaneously, the raw input is predicted into a *snow covered class label* output ScL. Finally, the intersection of both the outputs is the dangerous region of interest (ROI) *snow-over-road*. The ratio index based on two counting pixels of ROIs is computed by dividing the snow-over-road ROI by the road surface class ROI.

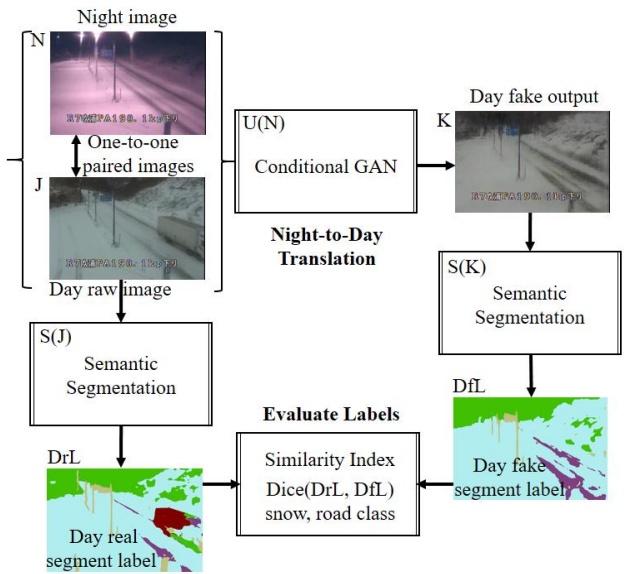

Figure 2: Overview of our pipeline showing the night image N is translated by the conditional GAN network U. The resulting day image fake output K is further fed into the semantic segmentation network S to predict a *day fake label* output DfL. Simultaneously, the day real input is predicted into a *day real label* output DrL. The similarity Dice index based on two labels is computed focusing the region of interest classes such as snow and road.

## 2. Night-to-Day Conditional GAN Translator

We propose a translator using the conditional GAN, and evaluate the snow-road classes label similarity between the day real segmented label and day fake segmented label. We formulate each task as follows.

### 2.1 Pix2pix for Translating Snowy Night-to-Day

The image-to-image translation is possible for training a paired image dataset with fixed camera angle. This study proposes a method to generate the road surface from snow-covered region using L1-Conditional GAN. In the original pix2pix study, the input edge images were translated to shoe images [Isola 2017]. Further, using the street photo dataset, they translated the day to night. However, in the case of snowy night road images, it is difficult to detect the snowy road surface. This study applies a generative deep learning method, in which the snowy road surface region ineffective outlook at night is translated to fake day output using the conditional GAN, pix2pix.

### 2.2 Semantic Segmentation for Snow Class

There are useful architectures such as semantic segmentation algorithms using deep learning for image analysis. The first end-to-end semantic segmentation neural network is the fully convolutional network (FCN). Several modifications have been made to FCN to improve performance. SegNet refines the encoder-decoder network layers with skip connections and conv-batchnorm-relu layers. DeepLab replaced the deconvolutional layers with densely convolution and atrous spatial pyramid pooling instead of standard convolution layers [Chen 2018].

This study applies a semantic segmentation method to detect snow-covered and road surface classes using a DeepLabv3+ algorithm with MobileNet as a backbone. This network has low memory and is practically accurate and fast for live camera monitoring of winter roads. We demonstrate this semantic segmentation network based on the six class defined labels such as road, pole-sign, green, snow, sky, and background.

### 2.3 Similarity Index between Real and Fake Output

As depicted in Figure 2, the night raw image is abbreviated as N. Based on the two trained networks such as a night-to-day conditional GAN network U and a semantic segmentation network S. We propose a pipeline to automatically translate from the night real image to day fake output. First, a raw night image is translated into a day fake output K = U(N) using the trained pix2pix network T. In addition, the day fake output K is indirectly detected into the *snow and road class label* output DfL using the trained semantic segmentation network S. Though we have prepared the one-to-one paired images, the day real image J is also segmented into the snow and road class label DrL using the trained segmentation network S. Finally, we evaluate the similarity of two labels that contains the targeted snow, using the Dice similarity index abbreviated as Dice(DrL, DfL). This similarity score is ranged on zero to one. The higher Dice score, the stronger similarity between the fake output and real label.

## 3. Applied Results

### 3.1 Night-to-Day Translation

Herein, we demonstrate that our pipeline could automatically compute a snow hazard ratio index at the cold region. Figure 3 depicts the 16 test images randomly selected. In contrast, Figure 4 illustrates the ground-truth images of the road surface without snow. Figure 5 depicts a road surface fake output translated from a raw input using a trained pix2pix.

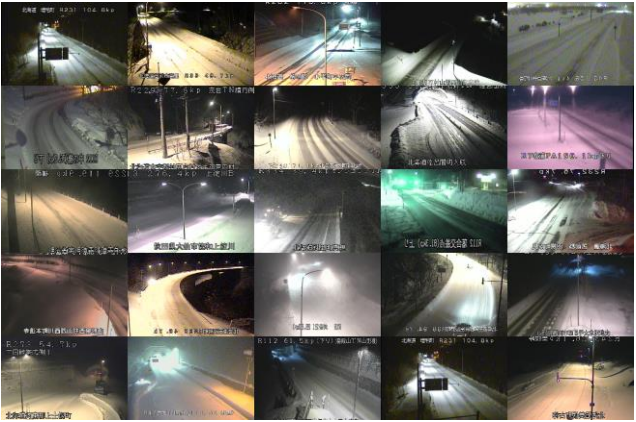

Figure 3: Night raw images examples (5 × 5 montage)

From [MLITT 2021] https://www.mlit.go.jp/road/road_e/index_e.html

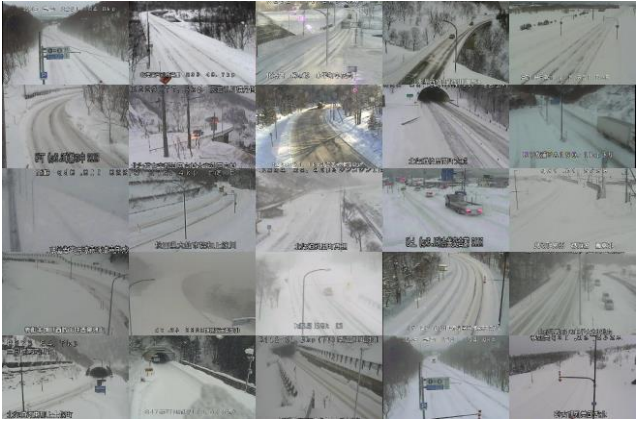

Figure 4: Ground-truth day images examples (5 × 5 montage)

From [MLITT 2021] https://www.mlit.go.jp/road/road_e/index_e.html

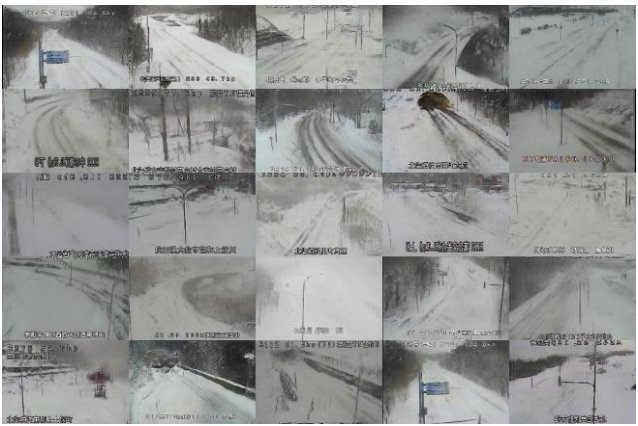

Figure 5: Fake day output translated from night raw images using pix2pix.

We prepared 1,130 one-to-one datasets that included two groups: the night snowy image and the day snowy image. We collected live road images from January 19 to 21 2021 at the cold region such as Yamagata, Akita, and Hokkaido in Japan [19]. We unified 480 × 720-size images for training. The ratio of training data versus test data was 90:10. We set the input size of pix2pix network as 512 × 768. We iterated 200 epoch using the Adam optimizer; therefore, it took 66 hours.

### 3.2 Predicted Segment Labels of Snow and Others

We prepared 500 annotated datasets that contained the raw images and six class defined labels such as road, pole-sign, green, snow, sky, and background. We resized images of 598 × 1196 size and extracted 2 × 4 crops, each with size 299 × 299; therefore, the total number of datasets was 4,000. The ratio of training data versus test data was 93:7. We set the input size of a semantic segmentation as 299 × 299. We trained 30 epoch iterations with a mini-batch of 16 and learning rate of 0.001 using the Adam optimizer; therefore, it took 3.5 hours. As presented in Fig. 5, each class mIoU had almost high accuracy scores, with the snow and road classes having the same score of more than 0.8. The entire accuracy mIoU was 0.7718, mean accuracy was 0.8572, and mean F1 score was 0.6886.

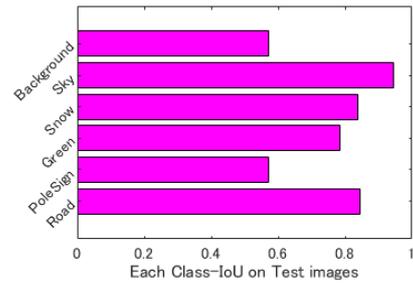

Figure 6: Each class mean intersection of union (mIoU) results.

As displayed in Figure 7, the day real output was segmented into the snow and road class label output using the trained network DeepLabv3+ with MobileNet as a backbone. As depicted in Figure 8, the fake day output was segmented into the snow class label output using the trained deep network. Compared with above two figures, the snow ROI has been well classified per pixel each other. Note that the road ROI has some cases different from the fake output and real label, e.g., 8[th] image.

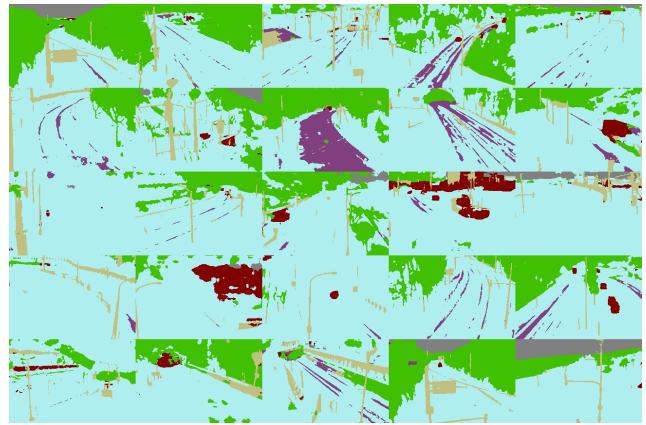

Figure 7: Day real label results predicted from the road surface fake outputs using DeepLabv3+ (5 × 5 montage)

As depicted in Figure 9, the Dice similarity index between the day fake output segmented label and the real day segmented label, has been high score respectively on the range from 0.75 to 0.95. Therefore, the fake day label output of snowy night image have been well approximated to the real label of snowy day per pixel as the critical ROI as the snow category for winter road safety.

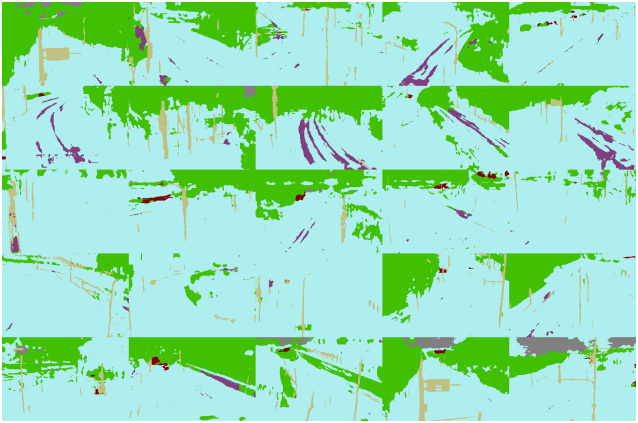

Figure 8: Fake day label results predicted from the input raw images using DeepLabv3+ (4 × 4 montage)

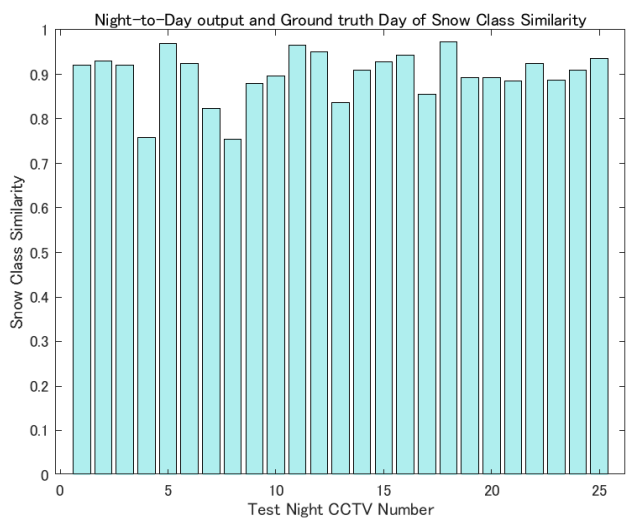

Figure 9: Bar plots of the similarity between two labels of the fake day output and real day image targeted ROI snow.

## 4. Concluding Remarks

This study proposed a snowy night-to-day conditional GAN translator using live CCTV images to monitor the winter road safety. The night snowy image has been translated to the fake day snowy image using a trained pix2pix. Further, we evaluated the similarity of targeted snow region of interest between the day real day label and the fake day output using a trained semantic segmentation. Finally, we collected live snowy road images from January 19 to 21 2021 at the cold regions such as Yamagata, Akita, and Hokkaido in Japan. Finally, we prepared the one-to-one paired images for the snowy night-to-day translator, and has been trained and built a practical translator. In result, the fake day label output of snowy night image have been well approximated to the real label of snowy day per pixel as the critical ROI as the snow category for monitoring the winter road safety.

Furthermore, we have already developed the previous pipeline to automatically compute the snow hazard indicator whose input is a snowy day image. This study have demonstrate a snowy night-to-day translation pipeline for monitoring the snowy night road, so as to approximate the night image to the day image. In future, we could combine above two pipeline for snowy night road status to make decisions on "road closures" and "snow removal work" owing to the road surface conditions even at night.

**Acknowledgments** I would like to thank Takuji Fukumoto and Shinichi Kuramoto (MathWorks) for providing us MATLAB resources for deep learning frameworks.